\documentclass{article}
\usepackage{spconf,amsmath,graphicx}

\usepackage{graphicx}
\usepackage[T1]{fontenc}
\usepackage{lmodern}
\usepackage{textcomp}
\usepackage{latexsym}
\usepackage{bm}  
\usepackage{color}
\usepackage{times,bm}
\usepackage{multicol}
\usepackage{epstopdf}
\usepackage{amsmath, amssymb}
\usepackage{type1cm}
\usepackage{amsmath,chemarrow}
\usepackage{cases}
\usepackage{pifont}
\usepackage{here}
\usepackage{float}
\usepackage[normalem]{ulem}
\usepackage{multirow}
\useunder{\uline}{\ul}{}

\title{
Unsupervised domain-adaptive person re-identification with multi-camera constraints
}
\name{Shun Takeuchi$^1$, Fei Li$^2$, Sho Iwasaki$^3$, Jiaqi Ning$^2$, Genta Suzuki$^1$
\thanks{\copyright\
2022 IEEE.  Personal use of this material is permitted. Permission from IEEE must be obtained for all other uses, in any current or future media, including reprinting/republishing this material for advertising or promotional purposes, creating new collective works, for resale or redistribution to servers or lists, or reuse of any copyrighted component of this work in other works.}
}
\address{
$^1$Fujitsu Research, $^2$Fujitsu R\&D Center, $^3$Fujitsu \\
{\small
\texttt{ \{takeuchi.shun, lifei, iwasaki.sho, ningjiaqi, suzuki.genta\}@fujitsu.com }
}
}

\begin{document}
%
\maketitle
\begin{abstract}
  Person re-identification is a key technology for analyzing video-based human behavior; however,
  its application is still challenging in practical situations due to the performance degradation for domains different from those in the training data.
  Here, we propose an environment-constrained adaptive network for reducing the domain gap.
  This network refines pseudo-labels estimated via a self-training scheme
  by imposing multi-camera constraints.
  The proposed method incorporates person-pair information without person identity labels obtained from the environment into the model training.
  In addition, we develop a method that appropriately selects a person from the pair that contributes to the performance improvement.
  We evaluate the performance of the network using public and private datasets and
  confirm the performance surpasses state-of-the-art methods in domains with overlapping camera views.
  To the best of our knowledge, this is the first study on domain-adaptive learning with multi-camera constraints that can be obtained in real environments.
\end{abstract}
%
\section{Introduction}
Person re-identification (ReID) aims to retrieve the same person (e.g., pedestrian) from different images.
It has attracted increasing attention in several industrial fields related to intelligent surveillance.
For example, in a retail situation, it is used as a base technology for analyzing consumers' in-store behavior, such as buying and shoplifting behaviors.

With the increasing availability of training datasets for ReID, several deep learning-based methods have been proposed (see~\cite{ye2021deep} and references therein).
However, its performance is limited to the same domain as the training data, and the performance significantly degrades for domains different from those in the training data~\cite{hu2014cross}.
This condition occurs because the characteristics of person images for the target domain differ from those for the source domain.
Because the annotation cost of the identity (ID) label for ReID is relatively higher than that for other image classification tasks, reducing the domain gap is required for its practical use.

Domain-adaptive learning methods using unlabeled data (UDA ReID: unsupervised domain-adaptive ReID) have been recently proposed~\cite{song2020unsupervised}. UDA ReID first trains a base model using labeled data, such as public datasets. Then, it uses the pre-trained model to infer the labels of target domain data.
The obtained labels (called pseudo-labels) are used for the target-domain model training.
However, inherently, the pseudo-labels of the UDA ReID framework contain noise, resulting in an insufficient performance.

\begin{figure} 
  \vspace{-2mm}
\centering
\centerline{\
\includegraphics[width=74mm]{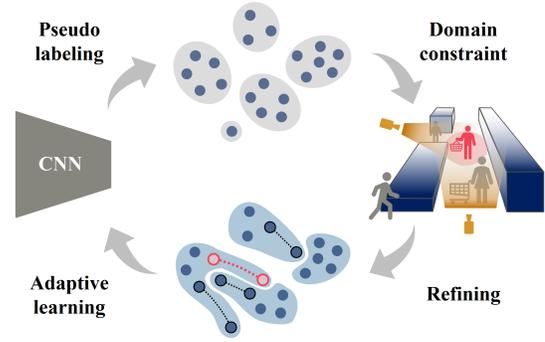}
}
\caption{
Illustration of the proposed method.
}
\label{fig:overview}
\vspace{-6mm}
\end{figure}

\begin{figure*} 
\centering
\centerline{\
\includegraphics[width=170mm]{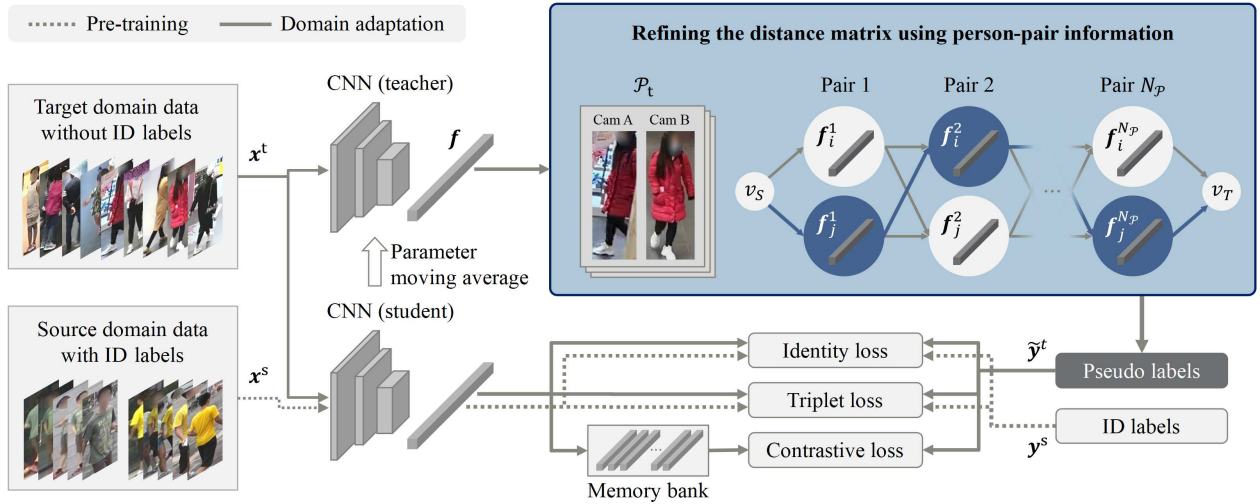}
}
\caption{
Framework of the proposed environment-constrained adaptive network.
}
\label{fig:proposed}
\end{figure*}

In this paper, we propose an environment-constrained adaptive network (ECA-Net) that alleviates the pseudo-label noise for the target domain (Figure~\ref{fig:overview}). The key feature of the proposed method is the use of information from a multi-camera environment. In retail industry, for example, a large number of surveillance cameras are installed in stores to analyze customer and employee behaviors. In such a situation, some overlapping regions occur in multiple camera views. Herein, we focused on acquiring the paired image data of the same person in the target domain from the overlapping region.
This approach does not require the ID label annotation and hence is promising for effectively acquiring target domain data.
However, note that the obtained paired images do not have person ID information.

ECA-Net is a revised UDA ReID method that incorporates the person-pair data without ID labels in the target domain and exhibits improved performance by alleviating pseudo-label noise.
Because there are no typical public ReID datasets with overlapping information,
we use public datasets annotated plausible overlapping information and a private dataset with overlapping information for the performance evaluation.
We confirm that the proposed method with person-pair images outperforms state-of-the-art methods.
Thus, it is a promising approach for improving the ReID's performance in real-world applications. The contributions of this paper are summarized as follows:
\begin{itemize}
  \item We propose a UDA ReID method (ECA-Net) with multi-camera constraints to reduce the domain gap.
  \item We develop an optimal selection strategy for the person from a person pair that contributes to the performance improvement.
  \item The performance surpasses state-of-the-art methods in domains with overlapping camera views.
\end{itemize}

\section{Proposed method}
\subsection{Overview}
Figure \ref{fig:proposed} shows an overview of the proposed ECA-Net method, which
is based on the mean teacher method, where the mean teacher model uses a temporally moving average of weights of the student network~\cite{tarvainen2017mean}.
The framework uses two datasets:
(1) source domain data, $\mathcal{D}_{\rm s} = \left\{\left(\bm{x}_i^{\rm s}, \bm{y}_i^{\rm s} \right) \, | \, i=1,\ldots, N_{\rm s} \right\}$;
(2) target domain data, $\mathcal{D}_{\rm t} = \left\{\bm{x}_i^{\rm t} \, | \, i=1, \ldots, N_{\rm t} \right\}$.
Here, $\bm{x}$ is the person image, $\bm{y}$ is the ID label, and $N$ is the number of samples.
First, the parameters of a convolutional neural network (CNN) are trained using $\mathcal{D}_{\rm s}$.
Next, using the obtained pre-trained model, the CNN parameters are optimized for $\mathcal{D}_{\rm t}$ through a self-training scheme.
The pseudo-labels are obtained by clustering the distance matrix data of each person.

We adopt ResNet-50 as the backbone of the CNN.
To improve the performance, the following techniques are used:
nonlocal block~\cite{wang2018non}, generalized-mean pooling~\cite{radenovic2018fine}, and batch normalization head~\cite{ioffe2015batch}.
In addition,
the ID classification loss, triplet loss, and contrastive loss are adopted as the loss functions~\cite{ye2021deep}.

\subsection{Domain adaptation}
On the UDA ReID framework, the distance between each person is calculated based on the CNN features $\bm{f}$ for each sample.
To calculate the distance matrix, we apply the widely used $k$-reciprocal re-ranking~\cite{zhong2017re}.
The $k$-reciprocal re-ranking uses the Jaccard distance, which is estimated based on the combination of neighboring images. It is given by
\begin{eqnarray}
d_{\rm{J}}\left(\bm{f}_i, \bm{f}_j \right) = 1 - \frac{\sum_{k=1}^N \min \left( M_{ik}, M_{jk} \right)}
{\sum_{k=1}^N \max \left( M_{ik}, M_{jk} \right)},
\label{eq:Jaccard}
\end{eqnarray}
where
\begin{eqnarray}  \label{eq:robustset}
 M_{ij} =
    \begin{cases}
        \exp \left\{ - d\left(\bm{f}_i, \bm{f}_{j} \right) \right\}   &   \text{if $j \in R(i, k)$}  \\
        0  &   \text{otherwise},
    \end{cases}
\end{eqnarray}
$d$ is the cosine distance between two CNN features, and
$R (i, k)$ is the $k$-reciprocal nearest neighbors of a sample $i$.
The pseudo-labels are extracted by performing clustering on the obtained distance matrix data,
$\tilde{\bm{y}}^t = F_{\rm{clust}} (d_{\rm{J}})$.

In the proposed method, the same person-pair list in the target domain is used to improve the performance for the target domain,
\begin{eqnarray}
  \mathcal{P}_{\rm t} = \left \{\left(\bm{x}_i^{\rm t}, \bm{x}_j^{\rm t} \right) \, | \, i \neq j \right \}.
\label{eq:scale}
\end{eqnarray}
Note that $\mathcal{P}_{\rm t}$ does not have the ID labels $\bm{y}^{\rm t}$.
The list denotes a set of person-pair images in each video frame
and the identities between different pairs are not given.
The ID labels are essential for training a high-performance ReID model~\cite{ye2021deep}.
Thus,
an approach that incorporates this incomplete data without the ID labels into the model is required.

We propose an approach to refine the distance matrix using the same person-pair list $\mathcal{P}_{\rm t}$.
As shown in Equations (\ref{eq:Jaccard}) and (\ref{eq:robustset}), the input variable for distance calculation is the CNN feature $\bm{f}$.
The closer the cosine distance of the CNN feature pairs, the closer is the Jaccard distance.
Using this property, we impose the following constraint,
\begin{eqnarray}
  \bm{f}_i = \bm{f}_j
\label{eq:constraint}
\end{eqnarray}
on the two-person pairs corresponding to $\mathcal{P}_{\rm t}$.
In conventional UDA ReID methods, when the image characteristics are different even for the same person, the Jaccard distance becomes large, and the same pseudo-label is not obtained.
In contrast, the proposed method can be guided to appropriate pseudo-labels by modifying the values of the CNN features
based on the same person-pair list obtained from the multi-camera environment.
Thus, based on the training data with the refined pseudo-labels,
$\mathcal{D}_{\rm t} = \left\{\left(\bm{x}_i^{\rm t}, \tilde{\bm{y}}_i^{\rm t} \right) \, | \, i=1,\ldots, N_{\rm t} \right\}$,
the model adapted to the target domain is trained.

\subsection{Optimal feature selection}
Equation~(\ref{eq:constraint}) indicates that images excluded as noise in the conventional methods are incorporated into the same pseudo-label.
Although we need to select the CNN features with a high confidence from each person pair, it is not obvious which feature is more appropriate.
Here, we focus on the property that
the number of $k$-reciprocal nearest neighbors $R$ for a noisy image is likely to be less than that for a noiseless image.
To incorporate more the noisy images into the noiseless images,
it is desirable to select the feature set with the largest sum of the number of $R$.
However,
because $R$ depend on the CNN features of all samples,
the possible number of combinations of $R$ for the number of pairs $N_{\mathcal{P}}$ becomes $2^{N_{\mathcal{P}}}$,
making enumerative searches impractical.

We develop a method that efficiently selects the optimal CNN features.
As shown in Figure~\ref{fig:proposed},
we define a directed graph $G = (V, E)$, denoted by edges $e_{ij} = (v_i, v_j) \in E$, which represents the selection of one image from a pair.
We then formulate it as a combinatorial optimization problem to maximize the sum of the number of $R$ in a path from the starting node $v_{\rm S} \in V$ to the terminal node $v_{\rm T} \in V$:
\begin{eqnarray}
\begin{aligned}
& \underset{z_{ij}}{\text{maximize}} && \sum_{e_{ij} \in E} a_{ij} z_{ij}              \\
& \text{subject to}                  && \sum_{v_i \in V} z_{hi} - \sum_{v_j \in V} z_{jh} =
    \begin{cases}
      1  &   \text{if $h = {\rm S}$}  \\
      -1 &   \text{if $h = {\rm T}$}  \\
      0  &   \text{otherwise},
    \end{cases} \\
&                                    && z_{ij} \in \{ 0, 1 \}, \; \forall e_{ij} \in E.
\end{aligned}
\label{eq:dijkstra}
\end{eqnarray}
The cost $a_{ij}$ is defined by
\begin{eqnarray}
  a_{ij} =
          \begin{cases}
              |R(j, k)| &   \text{if $j \neq {\rm T}$} \\
                      0 &   \text{if $j = {\rm T}$},
          \end{cases}
\label{eq:cost_function}
\end{eqnarray}
where
$R$ take into account the selection result in the previous nodes
and $|\cdot|$ denotes the number of candidates.
Dijkstra's algorithm can be applied to the proposed optimal feature selection,
and the time complexity becomes $\mathcal{O}(N_{\mathcal{P}}^2)$~\cite{dijkstra1959note}.
Thus, we efficiently estimate a plausible feature set based on $R$ that depend on some of the CNN features.

\section{Experiments}
\subsection{Datasets and evaluation metrics}
Three public datasets were used in the experiments:
(1)~Market-1501 (Market) contains 32,217 images from six cameras, where 751 identities were used for training and 750 for testing~\cite{zheng2015scalable};
(2)~DukeMTMC-reID (Duke) contains 36,411 images from eight cameras, where 702 identities were used for training and 702 for testing~\cite{zheng2017unlabeled};
(3)~MSMT17 (MSMT) contains 126,441 images from 15 cameras, where 1,041 identities were used for training and 3,060 for testing~\cite{wei2018person}.

The proposed method exploits the overlap between multiple camera views.
Although Market does not have any overlap information, some camera views are overlapped~\cite{zheng2017person}.
Based on the given information and video frame number, we acquired the same person-pair list $\mathcal{P}_{\rm t}$.
For Duke, we acquired the list based on the ratio of overlapping regions in the camera views~\cite{ristani2016performance}.
Because MSMT does not provide the full frame images of each camera,
we assumed that the person images before and after the camera switch on the frame time series overlap.
The ratio of the number of pair images and the amount of training data is 0.86 (Market), 0.007 (Duke), and 0.24 (MSMT).

Originally, the public datasets are not employed to evaluate the performance using the overlap information.
For this, we additionally evaluated the performance using a private dataset (Shopping mall).
It contains a video of pedestrians taken by three surveillance cameras installed inside a shopping mall in China.
The person's IDs were manually annotated, and the dataset contains 37,971 images, where 1,466 identities were used for training and 1,370 for testing.
A part of each camera view overlaps, and
the ratio of the number of pair images and the amount of training data is 0.13.

We adopted the mean average precision (mAP) (\%) and
cumulative matching characteristics (\%) of rank-1/5/10 for the performance evaluation.

\begin{table*}[h]
\vspace{-2mm}
 \begin{center}
 \caption{Performance compared with state-of-the-art methods. The best results are shown in boldface.}
 \label{table:comparison}
\begin{tabular}{lc|cccc|cccc}
\hline
              & \multicolumn{1}{c|}{}          & \multicolumn{4}{c|}{Duke $\to$ Market}  & \multicolumn{4}{c}{Market $\to$ Duke}  \\ \cline{3-10}
Methods        & \multicolumn{1}{c|}{Reference} & mAP        & Rank-1      & Rank-5      & Rank-10 & mAP        & Rank-1      & Rank-5      & Rank-10    \\ \hline

AD-Cluster \cite{zhai2020ad} & CVPR20 & 68.3 & 86.7 & 94.4 & 96.5 & 54.1 & 72.6 & 82.5 & 85.5 \\
MMT \cite{ge2019mutual} & ICLR20 & 71.2 & 87.7 & 94.9 & 96.9 & 65.1 & 78.0 & 88.8 & 92.5 \\
NRMT \cite{zhao2020unsupervised} & ECCV20 & 71.7 & 87.8 & 94.6 & 96.5 & 62.2 & 77.8 & 86.9 & 89.5 \\
MEB-Net \cite{zhai2020multiple} & ECCV20 & 76.0 & 89.9 & 96.0 & 97.5 & 66.1 & 79.6 & 88.3 & 92.2 \\
GLT \cite{zheng2021group} & CVPR21 & 79.5 & 92.2 & 96.5 & 97.8 & \textbf{69.2} & 82.0 & 90.2 & 92.8 \\
UNRN \cite{zheng2021exploiting} & AAAI21 & 78.1 & 91.9 & 96.1 & 97.8 & 69.1 & 82.0 & \textbf{90.7} & \textbf{93.5} \\

ECA-Net (Ours) & - & \textbf{83.4} & \textbf{94.0} & \textbf{97.5} & \textbf{98.3} & 68.3 & \textbf{82.3} & 90.4 & 92.5 \\ \hline \hline

              &                                & \multicolumn{4}{c|}{Market $\to$ MSMT} & \multicolumn{4}{c}{Duke $\to$ MSMT} \\ \cline{3-10}
Methods        & \multicolumn{1}{c|}{Reference} & mAP        & Rank-1      & Rank-5      & Rank-10 & mAP        & Rank-1      & Rank-5      & Rank-10    \\ \hline

NRMT \cite{zhao2020unsupervised} & ECCV20 & 19.8 & 43.7 & 56.5 & 62.2 & 20.6 & 45.2 & 57.8 & 63.3 \\
ABMT \cite{chen2021enhancing} & WACV21 & 23.2 & 49.2 & - & - & 26.5 & 54.3 & - & - \\
GLT \cite{zheng2021group} & CVPR21 & 26.5 & 56.6 & 67.5 & 72.0 & 27.7 & 59.5 & 70.1 & 74.2 \\
UNRN \cite{zheng2021exploiting} & AAAI21 & 25.3 & 52.4 & 64.7 & 69.7 & 26.2 & 54.9 & 67.3 & 70.6 \\

ECA-Net (Ours) & - & \textbf{35.0} & \textbf{65.1} & \textbf{76.1} & \textbf{80.2} & \textbf{36.6} & \textbf{66.9} & \textbf{78.0} & \textbf{82.0} \\ \hline \hline

&                                & \multicolumn{4}{c|}{Market $\to$ Shopping mall} & \multicolumn{4}{c}{Duke $\to$ Shopping mall} \\ \cline{3-10}
Methods        & \multicolumn{1}{c|}{Reference} & mAP        & Rank-1      & Rank-5      & Rank-10 & mAP        & Rank-1      & Rank-5      & Rank-10    \\ \hline

UNRN \cite{zheng2021exploiting} & - & 45.6 & 55.1 & 72.4 & 78.0 & 42.4 & 53.3 & 69.7 & 74.9  \\

ECA-Net (Ours) & - & \textbf{57.4} & \textbf{67.8} & \textbf{79.8} & \textbf{83.7} & \textbf{56.6} & \textbf{67.5} & \textbf{81.5} & \textbf{85.5}  \\ \hline

\end{tabular}
\end{center}
\vspace{-5mm}
\end{table*}

%
%
%
%

\begin{table}
\vspace{-3mm}
  \setlength{\tabcolsep}{1.3mm} 
  \begin{center}
  \caption{Comparison of different feature selection methods.}
  \label{table:selection}
\begin{tabular}{lc|cc|cc}
\hline
                         &                   & \multicolumn{2}{c|}{Duke $\to$ Market} & \multicolumn{2}{c}{Market $\to$ Duke}  \\
\cline{3-6}
Methods                  & Selection         & mAP           & Rank-1        & mAP           & Rank-1            \\ \hline
\multicolumn{2}{l|}{UNRN \cite{zheng2021exploiting} \;\;\;\; - }   & 78.1          & 91.9          & 69.1          & 82.0              \\ \hline
\multirow{3}{*}{\begin{tabular}[c]{@{}l@{}}ECA-Net\\(Ours)\end{tabular}} & Random            & 82.4 & 93.2 & 66.6 & 81.9              \\
                         & Partial           & 82.8          & 93.6          & 67.2          & \textbf{82.3} \\
                         & Optimal           & \textbf{83.4} & \textbf{94.0} & \textbf{68.3} & \textbf{82.3} \\ \hline \hline
                         &                   & \multicolumn{2}{c|}{Market $\to$ MSMT} & \multicolumn{2}{c}{Duke $\to$ MSMT}  \\
\cline{3-6}
Methods                  & Selection         & mAP           & Rank-1        & mAP           & Rank-1            \\ \hline
\multicolumn{2}{l|}{UNRN \cite{zheng2021exploiting} \;\;\;\; - }   & 25.3          & 52.4          & 26.2          & 54.9              \\ \hline
\multirow{3}{*}{\begin{tabular}[c]{@{}l@{}}ECA-Net\\(Ours)\end{tabular}} & Random            & 31.5 & 60.6 & 33.7 & 64.0 \\
                         & Partial           & 33.9          & 63.8          & 36.1          & 66.3 \\
                         & Optimal           & \textbf{35.0} & \textbf{65.1} & \textbf{36.6} & \textbf{66.9} \\ \hline
\end{tabular}
\end{center}
\vspace{-5mm}
\end{table}

\subsection{Implementation details}
For data preprocessing,
we performed data augmentation via random erasing, random cropping, and flipping.
Then, Adam optimizer was applied~\cite{kingma2014adam}.
The warm-up learning rate was used to prevent the overfitting of the CNN parameters~\cite{luo2019bag}.
DBSCAN was used as the clustering algorithm for estimating pseudo-labels~\cite{ester1996density}.
It does not require the specification of the number of clusters in advance
and is widely used in the conventional UDA ReID methods.

\subsection{Comparison with state-of-the-art methods}
We compared ECA-Net with other state-of-the-art methods in Table~\ref{table:comparison}.
ABMT~\cite{chen2021enhancing} does not provide the results of rank-5/10, and
the performances of UNRN~\cite{zheng2021exploiting} for Market $\to$ Shopping mall and Duke $\to$ Shopping mall
were evaluated independently using the official code.
The proposed method achieves the best performance for Duke $\to$ Market, Market $\to$ MSMT, Duke $\to$ MSMT,
Market $\to$ Shopping mall, and Duke $\to$ Shopping mall
and a comparable performance for Market $\to$ Duke.
The results indicate that the performance roughly correlates with the ratio of the number of pair images and the amount of training data.
Although the performance improvement for Duke is limited because the overlap of camera views was slight,
a sufficient performance improvement is achieved for Market, MSMT, and Shopping mall with several overlapping views.
In particular, for MSMT and Shopping mall with complex scenes,
ECA-Net significantly outperformed the existing methods.

Table~\ref{table:selection} shows the performance of the proposed optimal feature selection (Optimal) and other simple methods.
Random is a method to randomly select a person from the pair
and
Partial is a method to select the person with a large number of $k$-reciprocal nearest neighbors $|R|$ for each pair.
We confirm that the proposed method selects appropriate a CNN feature from the pair.
The $|R|$-based methods are effective for the feature selection.
Furthermore,
Optimal is more plausible than Partial because it is more desirable to have a larger sum of $|R|$ to increase the noiseless images.

\section{Conclusions}
In this paper, we have proposed a person ReID method (ECA-Net) for reducing the domain gap.
To the best of our knowledge, this is the first work on domain-adaptive learning with multi-camera constraints.
The proposed method incorporates obtained pair images without the ID labels into the training data.
It exhibits a performance superior to that of the existing state-of-the-art methods.
Because the acquisition cost of the pair images is relatively low compared to that of the person IDs,
the proposed ECA-Net would be a promising approach to improve the ReID's performance in practical situations.

\bibliographystyle{IEEEbib}

\end{document}